\documentclass{InsightArticle}

\usepackage[utf8]{inputenc}
\usepackage{graphicx}
\usepackage{subcaption}
\usepackage{booktabs}
\usepackage{textcomp}
\usepackage{longtable}

\usepackage[bookmarks,
bookmarksopen,
backref,
colorlinks,linkcolor={blue},citecolor={blue},urlcolor={blue}
]{hyperref}

\title{Scalable Simple Linear Iterative Clustering (SSLIC) Using a
  Generic and Parallel Approach}

\newcommand{\IJhandlerIDnumber}{3596}

\release{1.0}

\author{Bradley C. Lowekamp$^{1,2}$ and David T. Chen$^{1,2}$ and Ziv Yaniv$^{1,3}$ and Terry S. Yoo$^{1}$}
\authoraddress{$^{1}$National Library Of Medicine $^{2}$Medical Science and Computing LLC$^{3}$ TAJ Inc.}

\begin{document}

%
% Add hyperlink to the web location and license of the paper.
% The argument of this command is the handler identifier given
% by the Insight Journal to this paper.
%
\IJhandlefooter{\IJhandlerIDnumber}

\ifpdf
\else
   %
   % Commands for including Graphics when using latex
   %
   \DeclareGraphicsExtensions{.eps,.jpg,.gif,.tiff,.bmp,.png}
   \DeclareGraphicsRule{.jpg}{eps}{.jpg.bb}{`convert #1 eps:-}
   \DeclareGraphicsRule{.gif}{eps}{.gif.bb}{`convert #1 eps:-}
   \DeclareGraphicsRule{.tiff}{eps}{.tiff.bb}{`convert #1 eps:-}
   \DeclareGraphicsRule{.bmp}{eps}{.bmp.bb}{`convert #1 eps:-}
   \DeclareGraphicsRule{.png}{eps}{.png.bb}{`convert #1 eps:-}
\fi

\maketitle

\ifhtml
\chapter*{Front Matter\label{front}}
\fi

% The abstract should be a paragraph or two long, and describe the
% scope of the document.
\begin{abstract}
\noindent
Superpixel algorithms have proven to be a useful initial step for
segmentation and subsequent processing of images, reducing computational complexity 
by replacing the use of expensive per-pixel primitives with a higher-level abstraction, superpixels. 
They have been successfully applied both in the context of traditional image analysis and deep learning
based approaches. In this work, we present a generalized implementation of the simple linear iterative clustering
(SLIC) superpixel algorithm that has been generalized for n-dimensional
scalar and multi-channel images. Additionally, the standard iterative implementation
is replaced by a parallel, multi-threaded one. We describe the implementation
details and analyze its scalability using a strong scaling formulation. Quantitative evaluation 
is performed using a 3D image, the Visible Human cryosection dataset, and a 2D image
from the same dataset. Results show  good scalability with runtime gains even when using
a large number of threads that exceeds the physical number of available cores (hyperthreading).
\end{abstract}

\IJhandlenote{\IJhandlerIDnumber}

\tableofcontents

\section{Introduction}

Pixels, or voxels in three dimensions, are the basic primitive of an image,
usually defining a rectilinear grid. Superpixels reduce the number of primitives
representing an image  by grouping pixels based on low level features,
properties such as color, texture and physical proximity. Originally introduced
in~\cite{ren2003} as a method for reducing the complexity of higher-level image analysis tasks,
they have been successfully used in many computer vision tasks such as object
detection, depth estimation, and segmentation. A large number of algorithms for
creating superpixels have been proposed in the literature, with a recent
comparative evaluation of 28 algorithms described
in~\cite{DBLP:journals/cviu/StutzHL18}. One of the more popular and successful
superpixel algorithms is the Simple Linear Iterative Clustering (SLIC)
algorithm~\cite{DBLP:journals/pami/AchantaSSLFS12, lucchi2012}.

The SLIC algorithm has been used both in the context of classical image analysis
algorithms and in the context of deep learning. Examples of using SLIC in the
context of graph based algorithms include segmentation of mitochondria in
electron microscopy volumes~\cite{lucchi2012}, classification of hyperspectral
images~\cite{jia2017}, segmentation of the prostate in MR~\cite{tian2016}, and
segmentation of the liver in CT~\cite{wu2016}. Examples of using the SLIC
algorithm in combination with deep learning include segmentation of the pancreas
in CT~\cite{farag2017}, general salient object detection in color
pictures~\cite{he2015}, hyperspectral image classification~\cite{shi2018},
detection of cell nuclei in digital histology slides ~\cite{sornapudi2018}, and
classification of epithelial and stromal regions in histopathology
images~\cite{xu2016}.

The National Library of Medicine's Insight Segmentation and Registration Toolkit
(ITK) includes a couple of segmentation algorithms that could be classified as
superpixel methods; they are the toboggan image filter, the classic watershed
image filter and the morphological watershed image
filter~\cite{Beare+Lehmann2006}. These filters are all related to the original
watershed segmentation algorithm, operate on the gradient magnitude and perform
region growing with seeds from the local gradient magnitude minima. These
methods are greedy algorithms and single threaded, therefore they are neither
scalable for large data nor is the whole vector space taken into consideration
for the superpixel grouping when the image is non-scalar.

Our contribution of a scalable version of the SLIC algorithm is motivated by
work with several types of large images with a variety of characteristics.
These types include focused ion-beam scanning electron microscopy (FIB-SEM)
which forms 3D volumes with a single channel (gray scale). Typical image sizes are
more than 4 Gb with continued demanded for increased resolution and larger
volumes. Another large image type of interest is whole slide histology
imaging. Histology images are generally 2D three channel (RGB) images with a size of
several ($\leq$10) Gb. Finally, we are also interested in working with 3D
multi-channel confocal microscopy images whose size is also on the order of several
($\leq$10) Gb.

In the rest of this paper we describe the original SLIC algorithm, our parallel
and multi-dimensional version of the algorithm, Scalable SLIC (SSLIC), and an
evaluation of our algorithm's scalability using both a large 53Gb 3D color image
and a comparatively small 24Mb 2D color image.

\section{The Simple Linear Iterative Clustering (SLIC) Algorithm}

Our superpixel implementation is based on the SLIC algorithm proposed by Achanta
et al. \cite{DBLP:journals/pami/AchantaSSLFS12}. The goals outlined for the SLIC
algorithm include the following desirable properties with respect to the
resulting superpixels and the computation process: 1) The natural boundaries of
the image should be preserved by the boundaries of the superpixels. 2)
Computations of the superpixels should be quick, have low memory requirements
and involve only a few parameters. 3) The generated superpixels should improve
the accuracy and speed of subsequent segmentation steps.

The SLIC algorithm can be viewed as a specialized and optimized variation of
k-means clustering where each pixel is mapped to a point whose
coordinates correspond to a concatenation of the pixel coordinates and the
channel values for that pixel. The original algorithm dealt with 2D color images
using the CIE-Lab color space. Thus each pixel was mapped to a five vector $[L, a, b, x,
y]$ with clustering performed in this 5D space. A user
specified property of the SLIC superpixels is the expected size of the super
pixel, denoted by $S$. This restricted size enables the reduction of the global search space of
classic k-means to a local neighborhood in the image domain of size $2S \times 2S$.

\subsection{Distances in the joint range-domain (intensity-geometry) space}

Defining an image as $I:\Omega \rightarrow \mathbf{r}$, a joint range-domain (intensity-geometry)
cluster center is represented as $C^\prime=[\Omega \mathbf{r}]^\mathsf{T}$. For the case of a 2D
image with a CIE-Lab color representation the cluster center is \mbox{$ C_k =[L, a, b,
  x, y]^\mathsf{T} $}. The distance between any pixel and a cluster center is
defined as $D=\sqrt{d_c^2+\left(\frac{d_s}{S}\right)^2m^2}$, where $d_c$ and $d_s$
are the Euclidean distance for the separate range and domain,
respectfully. $S$ is a normalizing constant which is the expected size
of a cluster and $m$ is a user specified weighting parameter. When $m$ is reduced the $d_c$
component becomes more dominant causing color to be the main criteria for cluster affinity while when
it is increased the spatial regularity of clusters is emphasized. For 2D images of CIE-Lab color,
a range of $m\in[1,40]$ is suggested with 10 being the default.

This distance metric can easily be extended to gray-scale images or
general multi-channel images by the 2-norm of $d_c$,
Similarly, $d_s$ can support n-dimensional images. It is
worth observing that the range of intensity values effects the weight of
$d_c$ vs. the $d_s$ components. The CIE-Lab color space has a range
of $L\in[0,100]$, $a\in[-86.185, 98.254]$ and $b\in[-107.863, 94.482]$,
which needs to be considered when working with normalized data or data
with a 16-bit integer range. Also note that the number of
components of either the color or dimension will also affect the
weighting of the metric.

From a practical standpoint, the outer most square root of $D$ is not necessary, as squared
values maintain their ordering based on the squared distance. Additionally, the fortuitous use of squared
Euclidean distances removes additional uses of square roots. This fact
results in an actual implementation of simply the sum of the squares
of the difference between the cluster center and the joint range-domain
representation with a constant:
\mbox{$D=(L_{C_k}-L_i)^2+(a_{C_k}-a_i)^2+(b_{C_k}-b_i)^2+((x_{C_k}-x_i)^2+(y_{C_k}-y_i)^2)\frac{m^2}{S^2}$}

\subsection{Algorithm Details}

The SLIC algorithm consists of the following three stages:

\begin{description}
\item[1. Initialization:] The cluster centers $C_k$ are initialized by
regularly sampling the domain $\Omega$ at fixed intervals. Each center
is then perturbed to the location and value of the lowest gradient
magnitude\footnote{This gradient only applies when dealing with single
channel images. When the image has multiple channels we use the Frobenius
norm of the Jacobian matrix ($\|J\|_F \equiv \sqrt{\Sigma_{i=1}^m\Sigma_{j=1}^nJ_{i,j}^2)} $.} in its $3 \times 3$ neighborhood. Next, a label image, $l$, is
initialized to an undefined label and a distance image, $d$, is
initialized to $\infty$.

\item[2. Iterate till termination criterion satsified:] \hfill  \\[-4mm]

\begin{description}

\item[Iterate over $C_k$:] \hfill  \\[-4mm]

\begin{description}
\item[Update label and distance images:]  For all pixels, $\mathbf{x}$, in a $[2S \times 2S]$ region around
$C_k$, compute  $D(C_k, [I(\mathbf{x}),\mathbf{x}])$. If this distance is less than $d(\mathbf{x})$ 
update $l(\mathbf{x})=k$ and $d(\mathbf{x})$.

\item[Update clusters:] For all labels, compute new cluster centers based on the
updated pixel labels, where the new center for cluster $k$  is the mean of $[I(\mathbf{x}),\mathbf{x}]$
where $l(\mathbf{x})$ is equal to $k$.

\end{description}

\end{description}

\item[Terminate iterations if:] Distance between previous and current cluster centers is below a threshold or
we have reached the maximal number of iterations.

\item[3. Spatial connectivity enforcement:] Connectivity is not enforced in the
above steps so the cluster may not be fully connected for all
components. This post processing step examines labeled connected
components not connected to their cluster center.  Such a connected component is relabeled
so that it is connected to the "nearest'' label, or if the component is
of sufficient size, it is assigned a new label.
\end{description}

\section{The Scalable Simple Linear Iterative Clustering (SSLIC) Algorithm}

Two key principles guiding ITK algorithm development are that: (a) algorithms
should be designed to work with n-dimensional images having an arbitrary number of channels per
pixel, and (b) algorithms should take advantage of modern hardware to parallelize
computations. Our proposed SSLIC algorithm follows both principles
in a manner which satisfies the goals of the original algorithm while focusing on
significantly improving its speed. In addition, SSLIC generalizes the original
algorithm to n-dimensional images with an arbitrary number of channels per
pixel. We next describe our approach to implementing the SLIC algorithm in a
parallel manner.

\subsection{Algorithm Details}

The SSLIC algorithm consists of the following three stages:

\begin{description}
\item[1. Initialization:] The cluster centers $C_k$ are initialized by
regularly sampling the domain $\Omega$ at fixed intervals. Then
all cluster centers are updated in parallel so that they are moved to 
the lowest gradient magnitude location in the $3 \times 3$ neighborhood
of their original locations. Next, a label image, $l$, is
initialized to an undefined label and a distance image, $d$, is
initialized to $\infty$.

\item[2. Iterate till termination criterion satsified:] \hfill  \\[-4mm]

\begin{description}
\item [Create non-overlapping regions which split the image and update label and distance
images in parallel:] Iterating over $C_k$, if the cluster's $[2S \times 2S \ldots 2S]$ nD neighborhood
intersects the region assigned to the thread, for all pixels, $\mathbf{x}$, in this intersection,
compute $D(C_k, [I(\mathbf{x}),\mathbf{x}])$. If this distance is less than
$d(\mathbf{x})$ update $l(\mathbf{x})=k$ and $d(\mathbf{x})$.

\item [Create non-overlapping regions which split the image and update $C_k$ using a map-reduce scheme:]
Map - in each region iterate over the pixels and accumulate the joint intensity-geometry information per
label. Reduce - merge the information from all regions based on the label ids and update the cluster centers
where the new center for cluster $k$  is the mean of the joint intensity-geometry information obtained for
label $k$ in the previous step.

\end{description}

\item[Terminate iterations if:] We have reached the maximal number of iterations (distance between previous and current
cluster centers is computed and available).

\item[3. Spatial connectivity enforcement:]  \hfill  \\[-4mm]
\begin{description}
\item[Initialize] a marker image $m$ to a value indicating that the label
at that location is not the final label. 
\item[In parallel] for each cluster center, if the label at the location defined by
$C_k$ is equal to $k$ (our cluster is not torus shaped), or we found the label $k$ in the   $[S \times S \ldots S]$ nD neighborhood
centered on $C_k$, obtain the connected component with label $k$ using this initial seed point. If this connected component's
size is greater than $\frac{S^n}{4}$, update the maker image in all these locations to indicate the label is final.
\item[Iterate over $m$], if $m(\mathbf{x})$ is not final, obtain the connected component with label $l(\mathbf{x}) $ using $\mathbf{x}$
as the seed point. If the size of this connected component is larger than $\frac{S^n}{4}$, change the label image for all these locations
to a new label, $k+1$, otherwise change it to the last encountered label and update the marker image to indicate that the label is final.

\end{description}
 
\end{description}

\subsection{SSLIC Parameters}

The SSLIC filter exposes two user adjustable parameters of interest: the desired
super grid size and the spatial weight factor which balances between superpixel spatial
regularity and color affinity.

The desired grid size in the original SLIC algorithm was a single number which
is appropriate for isotropic pixels. As our goal  is to accommodate images from
a variety of sources, many of which are highly un-isotropic, we allow the the
size of the superpixel to be specified as the number of pixels in each dimension
i.e. $[S_x, S_y, S_z]$. Therefore the superpixels themselves can be anisotropic
to accommodate non-uniform pixel spacing, as is common in medical images.

The weight factor is utilized to balance between the spatial and image intensity
portions of the distance metric. The default value is 10, which provides good
results for 2D images in the CIE-Lab color space. Increasing the value increases
the weight of the spatial component which produces more regularly shape and
sized superpixels. Image dimensionality, and similarly the magnitude of the
range of the pixels values will effect the relative weight between the two
components of the distance metric and may require experimentation to identify
the relevant weighting for a specific setting ($nD$ image with $c$ channels per
pixel).

Additionally, the user can specify the algorithm's termination criteria via the
maximal number of iterations, while the residuals or the change in cluster centers between two consecutive iterations can be monitored. The maximum number of iterations defaults to 5 for images with dimension 3 or greater, whereas the original SLIC implementation specifies 10 iterations for 2D images.

\subsection{Implementation Details}

Achieving the goals of memory and computational efficiency, while still supporting
grayscale, fixed vector and dynamic vector images requires some planning
and prudent choices for data-structures and memory layout. The resulting output
$l$ is clearly a label image, and the intermediate per pixel distance values,
$d$, are also represented by an image. Efficiently supporting variable length vectors in
ITK can be challenging due to the potential memory allocation per pixel to
support the run-time length. To avoid this, we store the set of cluster centers,
whose lengths are the number of dimensions plus the number of components for the
pixel value, in a single 1-dimensional array. The values of a cluster $C_k$ are
simple accessed via a 'vnl\_ref\_vector', which references the data in the array.

The user provided superpixel grid size specifies the expected size in pixel
units, not physical units as is common in ITK. The use of pixel units enables the grid
size parameter to be independent of the image spacing, removes potential
degenerate cases, allows reasonable default values, and follows that the
superpixels are an abstraction from the pixels. Therefore the "distance" metric
computed must be computed in index space and not physical space. We have also
extended the grid size to potentially be isotropic. So the spatial weights are
applied thusly: $D=\sqrt{d_c^2+\Sigma_{i}\left(\frac{d_i}{S_i}\right)^2m^2}$.

\section{SSLIC Evaluation}

As the focus of our algorithm was on improving the runtime of the original
SLIC algorithm without changing the original algorithmic approach we limit 
our evaluation to computational performance and scalability. 

In general, the time it takes to perform a task is comprised of the time it takes to
complete its sequential portion and the time it takes to perform its
parallel portion:
$$
T = T_s + T_p
$$

In our evaluation we use the concept of strong scalability. That is, the problem size
is kept fixed while we increase the number of parallel process (in our case these are lightweight threads).

The relative speedup\footnote{"Relative speedup" uses the single process implementation of the parallelized algorithm and not the best sequential algorithm which would correspond to "speedup".} obtained by using more than a single process is defined as:
$$
S(p) = \frac{T(1)}{T(p)},\;\; \textrm{where } T(1) \textrm{ is the runtime of the parallel implementation using a single process.}  
$$

The optimal relative speedup value is $S^*(p) = p$. 

The relative efficiency is defined as speedup divided by the number of processors:

$$
E(p) = \frac{S(p)}{p}
$$

The optimal relative efficiency is thus $E^*(p)=1$.

When evaluating using strong scalability we have an upper bound on the possible relative speedup and efficiency which are given
by Amdahl's law~\cite{amdahl1967}. Given that a fraction, $\alpha \in [0,1]$, of the task is serial we have:
$$
S(p) = \frac{T(1)}{\alpha T(1) + (1-\alpha) T(p)} \leq \frac{1}{\alpha + (1-\alpha)/p}
$$ 
 
and

$$
E(p) = \frac{T(1)}{p(\alpha T(1) + (1-\alpha) T(p))} \leq \frac{1}{1 + \alpha(p-1)}
$$ 

\begin{figure}
\centering
\begin{subfigure}{0.4\textwidth}
  \includegraphics[width=1.0\textwidth]{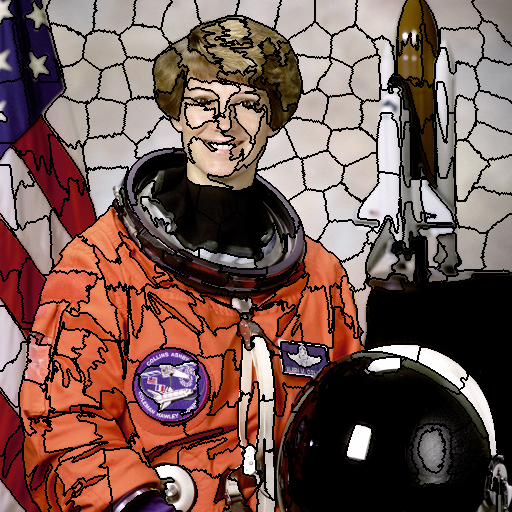}
  \caption{SSLIC implementation}
  \label{fig:a}
\end{subfigure}
\begin{subfigure}{0.4\textwidth}
  \includegraphics[width=1.0\textwidth]{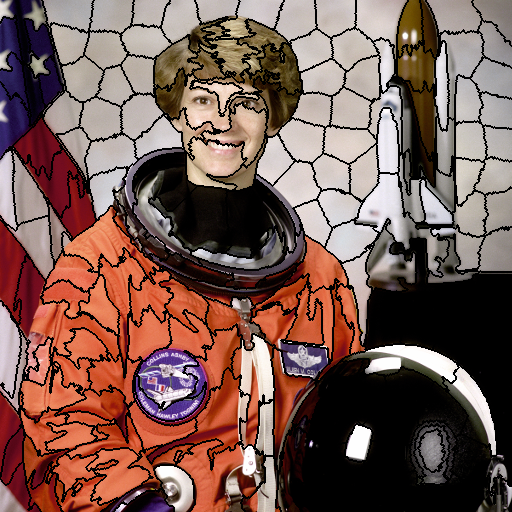}
  \caption{scikit-image SLIC implementation}
  \label{fig:b}
\end{subfigure}
\caption{Comparison of our SSLIC implementation and the scikit-image~\cite{scikit-image} SLIC
  implementation on a $512 \times 512$ color image converted to CIE-Lab color
  space. Each implementation was run for 10 iterations and the size of the
  superpixels was specified as $[32 \times 32]$ for our implementation, and a
  requested number of superpixels of 256 for scikit-image. The runtimes for our
  implementation were 292ms, 86ms, and 52ms with 1, 4 and 8 threads
  respectfully, while scikit-image's single threaded implementation runtime was 166ms.}
\label{fig:astronaut}
\end{figure}

\begin{figure}

\centering
\begin{subfigure}{0.67\textwidth}
  \includegraphics[width=1.0\textwidth]{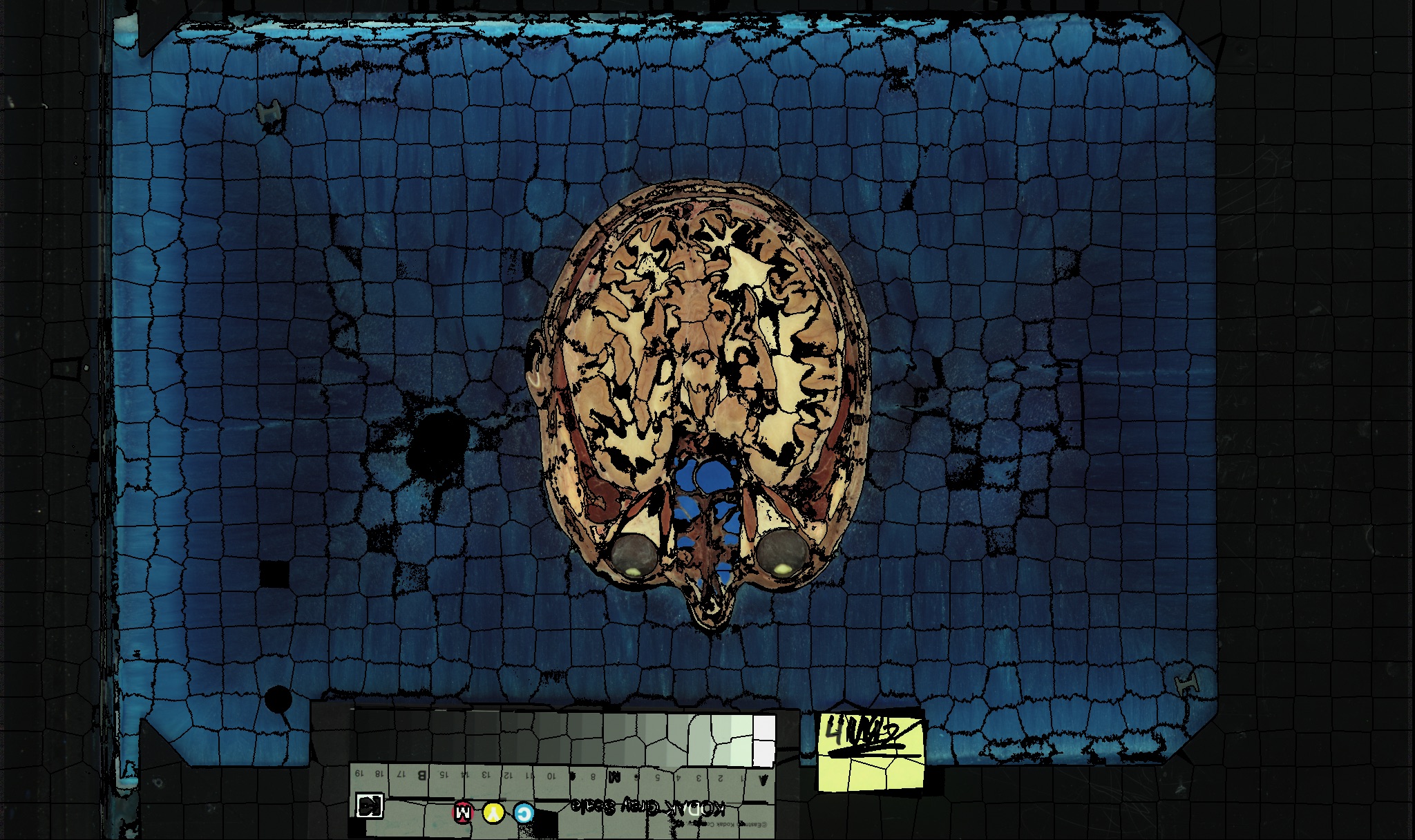}
%  \caption{a}
  \label{fig:a}
\end{subfigure}
\begin{subfigure}{0.67\textwidth}
  \includegraphics[width=1.0\textwidth]{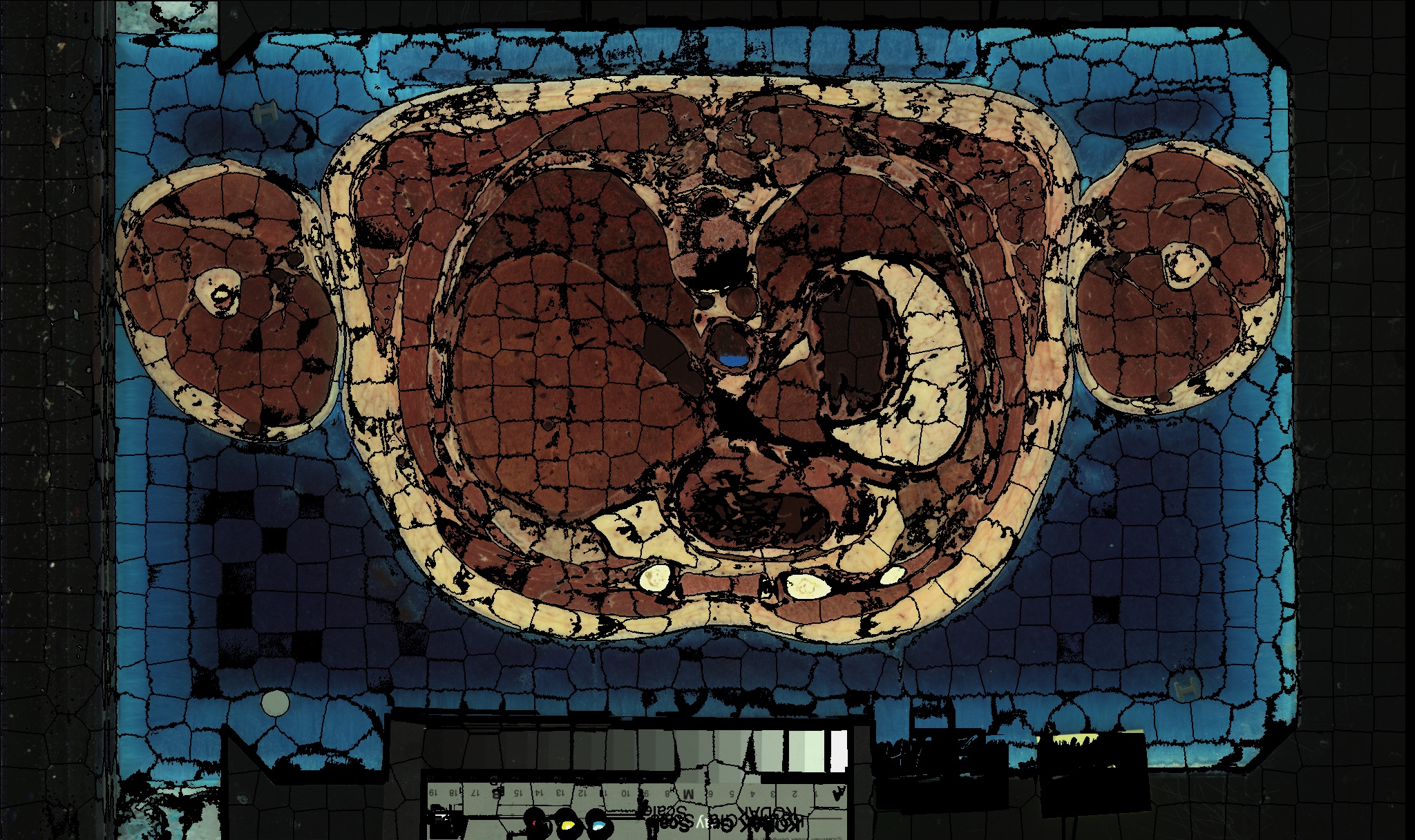}
%%  \caption{b}
  \label{fig:b}
\end{subfigure}
\begin{subfigure}{0.67\textwidth}
  \includegraphics[width=1.0\textwidth]{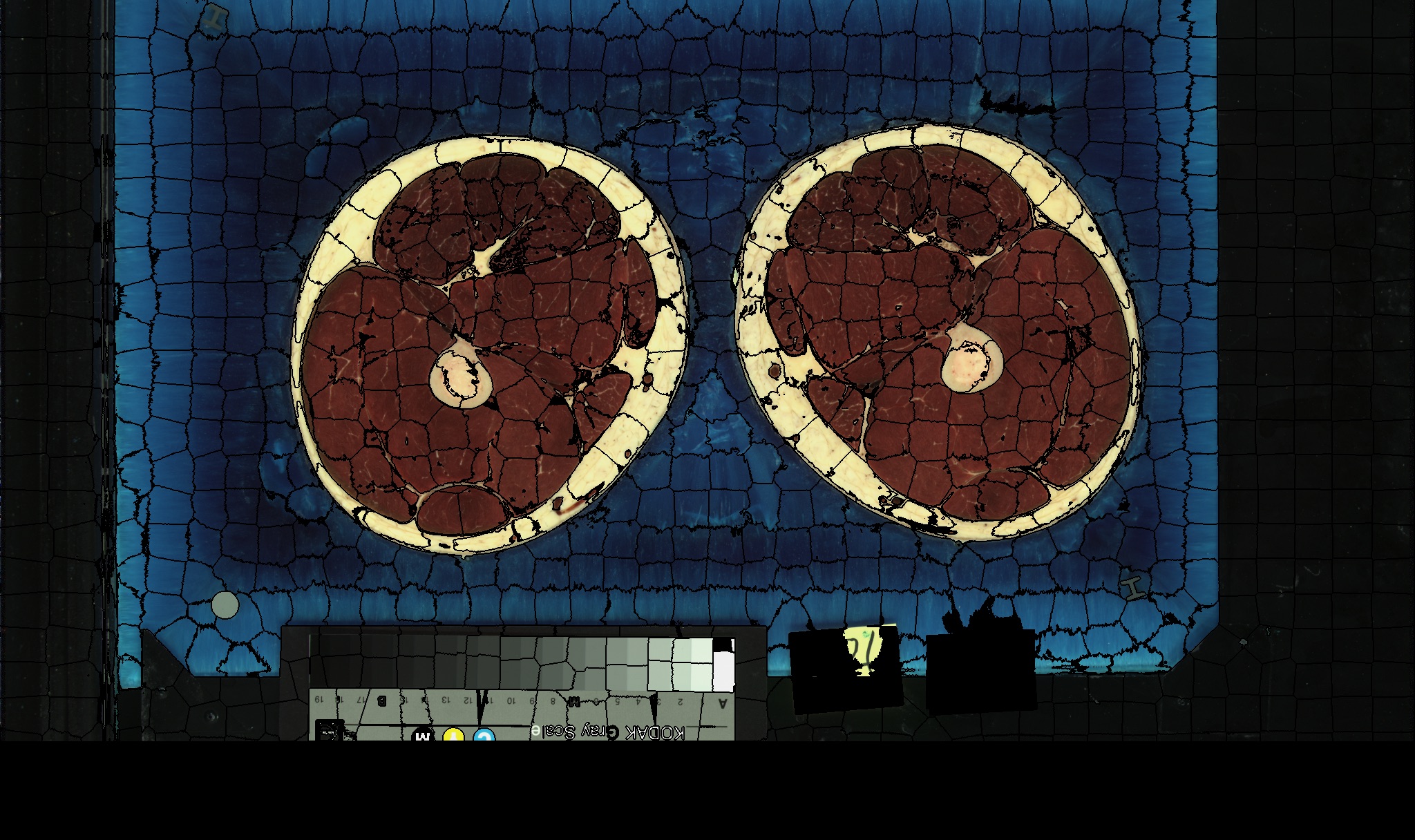}
%%  \caption{c}
  \label{fig:c}
\end{subfigure}
\caption{Selected slices of the Visible Human Male (110,487,116) with rendered superpixel borders. The SSLIC parameters were super grid $[30 \times 30 \times 10]$, spatial weight 10, and 5 iterations. The color slices were converted to CIE-Lab color space. The algorithm and superpixels are in 3D, so black regions may be caused by co-planar slice and superpixel boundaries. }
\label{fig:vh_slices}
\end{figure}

\begin{figure}

\centering
\begin{subfigure}{0.8\textwidth}
  \includegraphics[width=1.0\textwidth]{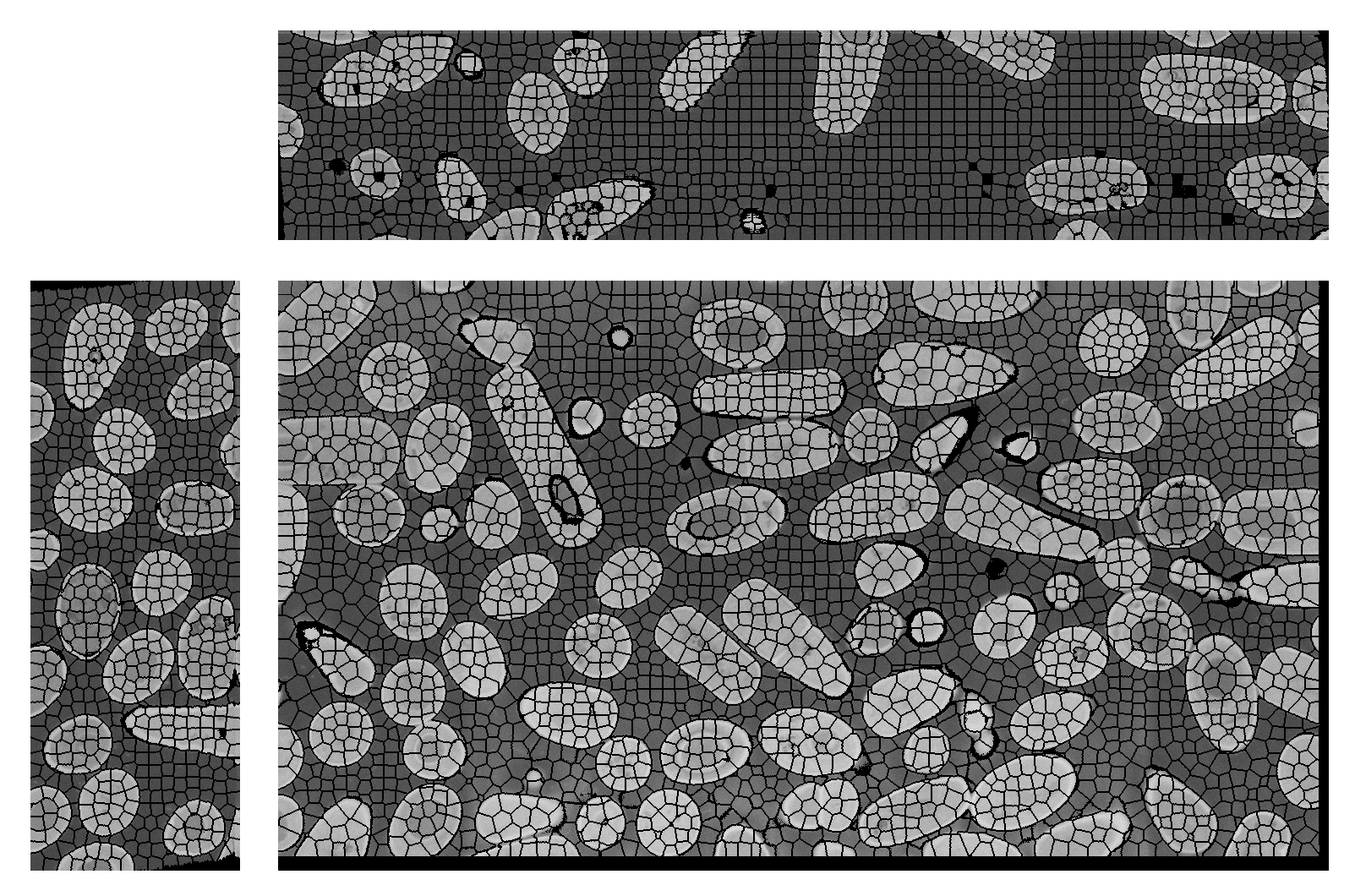}
\end{subfigure}
\caption{Planar cross sections of Bacillius subtilus acquired with focused
	ion-beam scanning electron microscopy. The SLIC parameters were super grid $[15
	\times 15 \times 15]$ spatial weight 5, and 10 iterations. From the $[ 1243
	\times 2094 \times 247]$ volume a selected X-Y planar slice is on the lower
	right, a Y-Z slice is on the left, and an X-Z slice is at the top.}
\label{fig:bsub_3d}
\end{figure}

\subsection{Method}

To evaluate the performance of the SSLIC algorithm we utilize the cyrosection
Visible Human Male dataset~\cite{DBLP:conf/medinfo/Ackerman98}. A frozen male
cadaver which was serially imaged and sectioned at 1 millimeter intervals to
form The color volume  Visible Human of $[2048 \times 1216 \times 1978]$ voxels.
The size of the original RGB (unsigned char) volume is 16Gb, and 53 Gb after
conversion to CIE-Lab (float), which is the data used in this work.
A single 2D slice is evaluated in addition to the whole volume to enable
performance comparison at two problem set sizes.

The computer system used for the performance analysis is a two CPU socket
server running Red Hat Enterprise Linux Server release 7.5. The CPUs are Intel 
Zeon CPU E5-2699 v4 @2.20GHz each having 22 physical cores and Hyper-Threading
enabled, resulting in 44 physical cores or 88 virtual cores. The
system has 512 Gigabytes of memory which is sufficient for processing the
dataset without swapping to disk.

To analyze the scalability of the SSLIC algorithm, the time of execution is
measured for a fixed image while varying the number of threads allocated to the
task. This was implemented in a python script via SimpleITK
bindings~\cite{lowekamp2013}. The reported timing is of the SimpleITK
\texttt{Execute} method which includes the construction and setting of ITK
parameters therefore this approach adds some negligible constant overhead when compared to directly
executing the ITK filter. The execution is timed with and without the
connectivity enforcement step. As this post-processing step involves a single
threaded pass through the entire image, the separate timings enables the
scalability assessment of the two algorithmic stages independently.

The code was built against the latest stable ITK release 4.13.0 with C++11
enabled for improved compatibility with the forth coming ITK 5.0 release. The
system compiler, "gcc (GCC) 4.8.5 20150623 (Red Hat 4.8.5-28)" was used with the
default ITK flags for "release" mode.

The method was first executed on the extracted 100th slice of the Visible
Human with dimensions of $[2048 \times 1216 ]$ pixels. The SSLIC algorithm ran
for 5 iterations with an isotropic supergrid size of 50, and the default spatial
weight of 10. This test case was executed 5 times, and the minimum time is
reported. With the brief execution time of less than a second, the number of
threads allocated to the SSLIC was incremented by 1.

Next the algorithm was evaluated on the whole 53 Gigabyte 3D Visible Human
dataset. The same parameters were specified: 5 iterations, 50 supergrid size,
and spatial weight 10. The algorithm was only run once for a selection of number
of threads.

Visualization of the resulting multi-label segmentation image is done with a
brief line of SimpleITK code (see below). Since the label ID or value of the
result contains no significant meaning, only the boundaries of the superpixels
are important, we follow the convention to render images using a
black contour around the segmentation.

\begin{verbatim}
def mask_label_contour(image, seg):
   """Combine an image and segmentation by masking the segmentation contour.

   For an input image (scalar or vector), and a multi-label
   segmentation image, creates an output image where the countour of
   each label masks the input image to black."""

   return sitk.Mask(image, sitk.LabelContour(seg+1)==0)

\end{verbatim}

\subsection{Results}

The following two section describe our qualitative and quantitative evaluation
of SSLIC. Superpixel labeled images are included for qualitative evaluation from
a selection of datasets to represent some of the diverse image types the SSLIC
algorithm is capable of operating upon. The quantitative sections focuses on
analyzing the performance and scalability characteristics of our implementation.

\subsubsection{Qualitative}

We demonstrate the results of our method on 3 distinct and representative
datasets. First is a photographic example of an astronaut in figure \ref{fig:astronaut}
from the scikit-image\cite{scikit-image} project. Visually, both implementations yield similar
results. In figure~\ref{fig:vh_slices}, three extracted slices from the 3D Visible Human dataset are shown
with 3D superpixels overlaid onto the slices. To capture the
anisotropic voxel size of $[0.3 \times 0.3 \times 1.0]mm$, the super grid size
was specified as $[30 \times 30 \times 10]$, with the default spacial proximity
of 10 and 5 iterations. The last dataset is a 3D focused ion-beam scanning
electron microscopy (FIB-SEM) of Bacillus subtilis bacterium courtesy of the High
Resolution Electron Microscopy at the National Cancer Institute, National
Institutes of Health \cite{Narayan2015}. The SSLIC algorithm was run on a
processed scalar volume of $[ 1243 \times 2094 \times 247]$ pixel with spacing
of approximately $[12 \times 12 \times 12]nm$, see figure \ref{fig:bsub_3d}

\begin{figure}
	\center
	\includegraphics[width=1.0\textwidth]{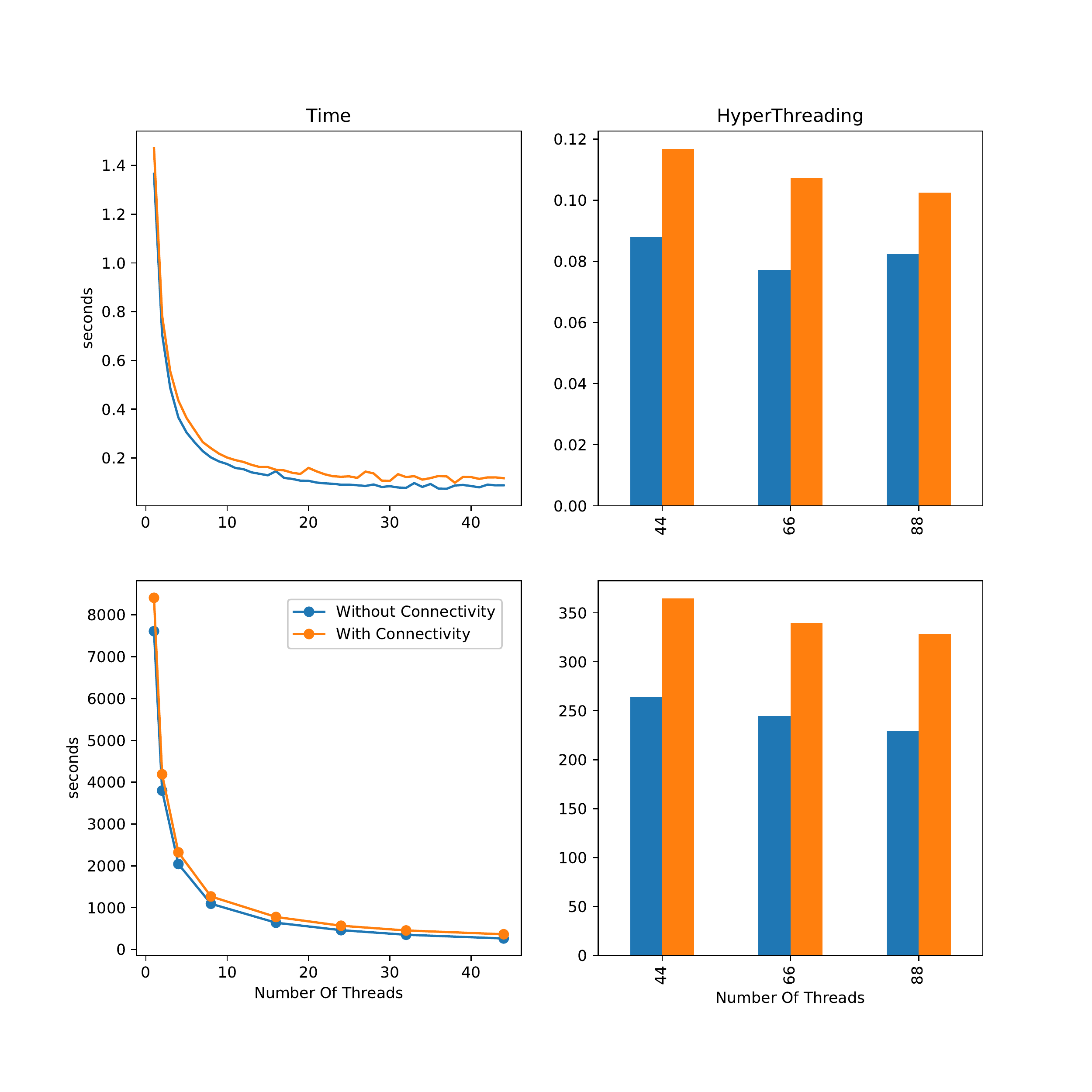}
	\itkcaption[Thread Timing]{Execution times of the ITK
	SSLIC filter with a varied number of threads demonstrating SSLIC's scalability.
	Top row: results obtained using a single 2D slice of the Visible Human Male with $[2048 \times 1216]$
	pixels. Bottom row: results obtained using the full 3D cyrosection dataset at $[2048 \times
	1216 \times 1978]$.} \label{fig:thread_time}
\end{figure}

\begin{figure}
	\center
	\includegraphics[width=1.0\textwidth]{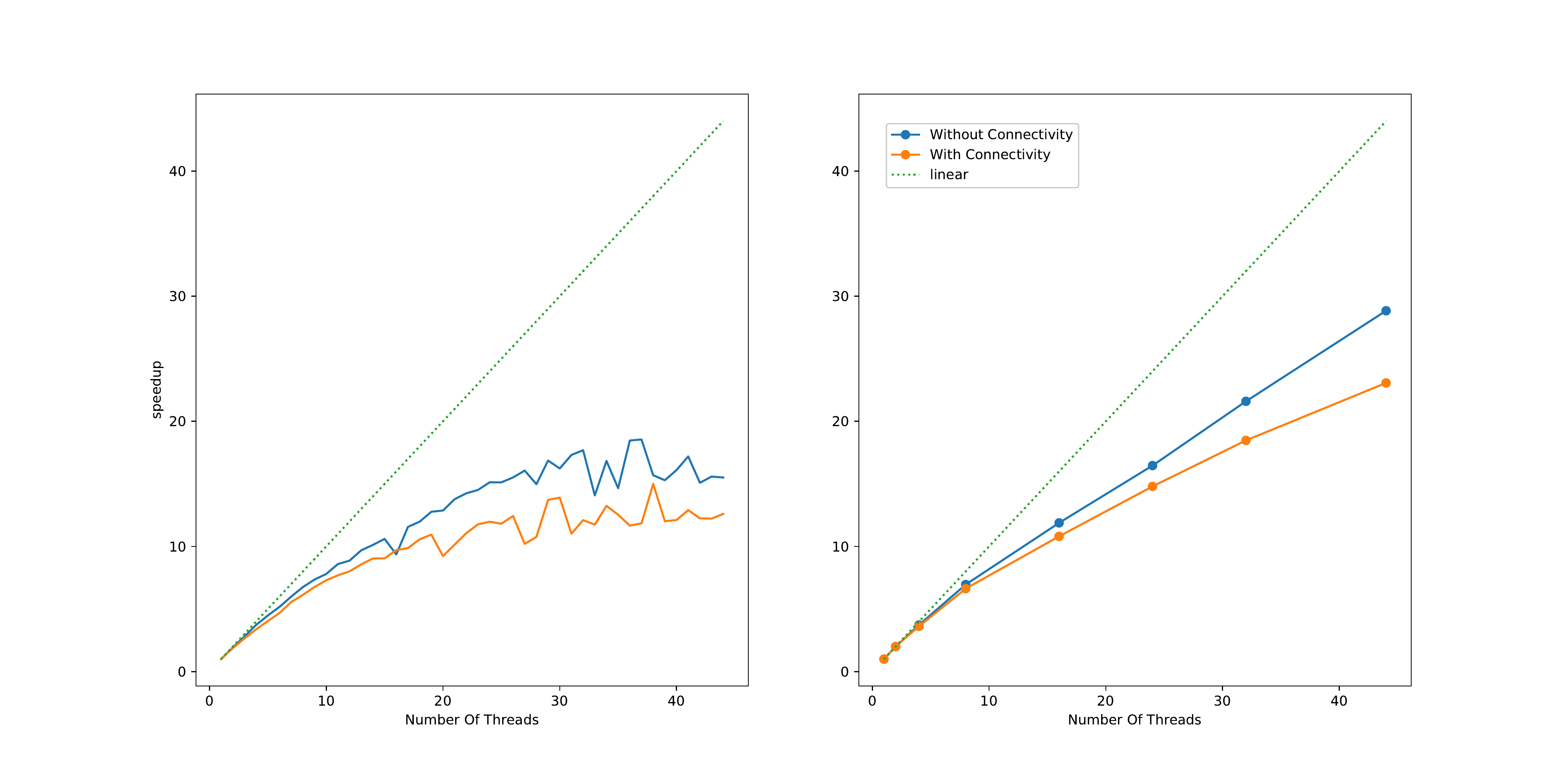}
	\itkcaption[Thread Speedup]{Speedup obtained by
		the ITK SSLIC filter with a varied number of threads. Comparison of
		the scalability on a 2D slice (left) and the 3D volume
		(right). The ideal linear speed up is a green dotted line.}
	\label{fig:thread_speedup}
\end{figure}

\subsubsection{Quantitative}

The detailed SSLIC performance timing results for the 2D
slice are given in table \ref{tab:vh_2d} and those for the 3D volume are in table
\ref{tab:vh_3d}. Included in the tables are the
computed relative efficiency and relative speedup as defined above.
These timing measurements are also summarized  in
figures~\ref{fig:thread_time} and~\ref{fig:thread_speedup}.

The 2D speedup graph shows the upper bound for speedup being approached
demonstrating Amdhal's law. That is to say for this relative small 2D image we
are bounded by the single threaded execution and overhead of the algorithm. This
is in contrast to the continued speedup for the 1978 times larger 3D dataset.
Performance gains continue when more resources are allocated to the problem. The
efficiency best quantifies the difference between 2D and 3D at 44 physical
cores, where the 2D case has 35\% while the 3D case has 66\% computed relative
efficiency. The phenomena of improved efficiency on larger datasets is described
by Gustafson's law\cite{Gustafson:1988}.

When the number of cores exceeds the number of physical cores, or when
HyperThreading is needed for virtual thread execution (although always enabled
on the system during evaluation), the results are separated into a bar graph in
figure \ref{fig:thread_time}. The HyperThreaded cores are a distinct type of
resource from a physical core as the virtual cores share many of the same CPU
physical resources such as cache and execution instructions with another. The
addition of virtual cores is not expected to provide  similar scalability
as additional physical cores. Despite low efficiency or utilization of the
virtual cores, our results demonstrate that utilizing HyperThreading yields
improved performance and decreased execution time even in the extreme case with
88 threads.

\section{Updating with Modern Threads}

The forthcoming ITK version 5.0 release includes a number of performance
enhancements to modernize the classic ITK threading model. The additions include
multiple threading back ends such as a thread-pool and an Intel Threading
Building Blocks (TBB) multi-threader interface. The latter supports dynamic load
balancing and advanced task scheduling. Our initial development of SSLIC
targeted the ITK version 4 interface. These emerging threading features were
considered during the initial SSLIC implementation which enabled updates to
support the new threading models.

The SSLIC implementation is updated to use the ITK version 5 threading model
while leaving the description of the parallelism identical. The implementation
changes from using thread barriers in the \texttt{ThreadedGenerateData} method
to a single threaded \texttt{GenerateData} method, which invokes the new
\texttt{ParallelizeArray} and \texttt{ParallelizeImageRegion} methods for each
parallel step as appropriate. Additionally, a mutex lock is introduced to
control access for the accumulation of the updated clusters. With this updated
implementation, thread identifiers and persistent per thread allocated storage
are removed from the multi-threaded step methods.

To evaluate the performance of this update the same 88 core system with the same
GCC 4.8.5 compiler applied to the same extracted 100th slice of the Visible Human with
dimensions of $[2048 \times 1216 ]$ pixels is analyzed. The ITK code at hash
`5470170e` is used for the original version of the SSLIC filter, and the
update to the modern ITKv5 threading model is applied. The SSLIC algorithm
timed performance is sampled 20 times, an increase to account for sensitivity in
the ratio used in the speedup formula. The reported time is the minimum of 20
executions for the algorithm. We run both the original version and the new ITKv5
implementation for the three supported multi-threaders: Intel TBB, thread pool,
and native platform (Table \ref{tab:ITKv5_time_wo}, \ref{tab:ITKv5_time_w}). To
quantify the performance difference of the new implementation we look at the
speedup of the ITKv5 update by the ratio of times: $original\ time/new\ time$.
Also the speed up of ITKv5 implementation compared to the classic platform
multi-threader is evaluated (Table \ref{tab:ITKv5_speedup_wo},
\ref{tab:ITKv5_speedup_w}).

\begin{figure}
	\center
	\includegraphics[width=1.0\textwidth]{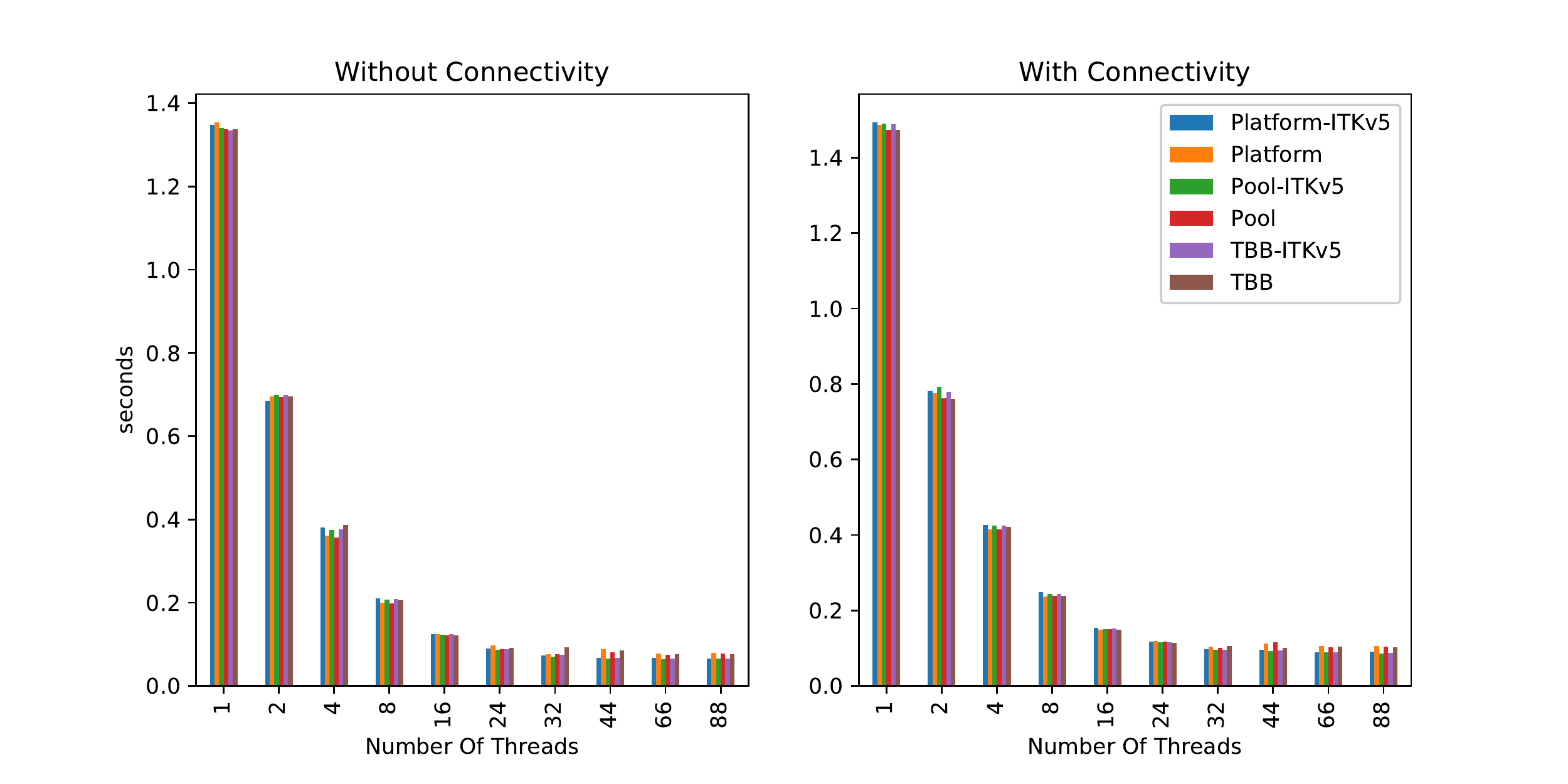}
	\itkcaption[Thread Timing]{Execution times of the SSLIC filter with a varied number of threads, the original ITKv4 implementation, the ITKv5 implementation, and varied with the different ITK multi-threaders using a single 2D slice of the Visible Human Male with $[2048 \times 1216]$
		left: without the connectivity step
		right: with the connectivity step that has a significant single threaded component. }
	\label{fig:ITKv5_time}
\end{figure}

\begin{figure}
	\center
	\includegraphics[width=1.0\textwidth]{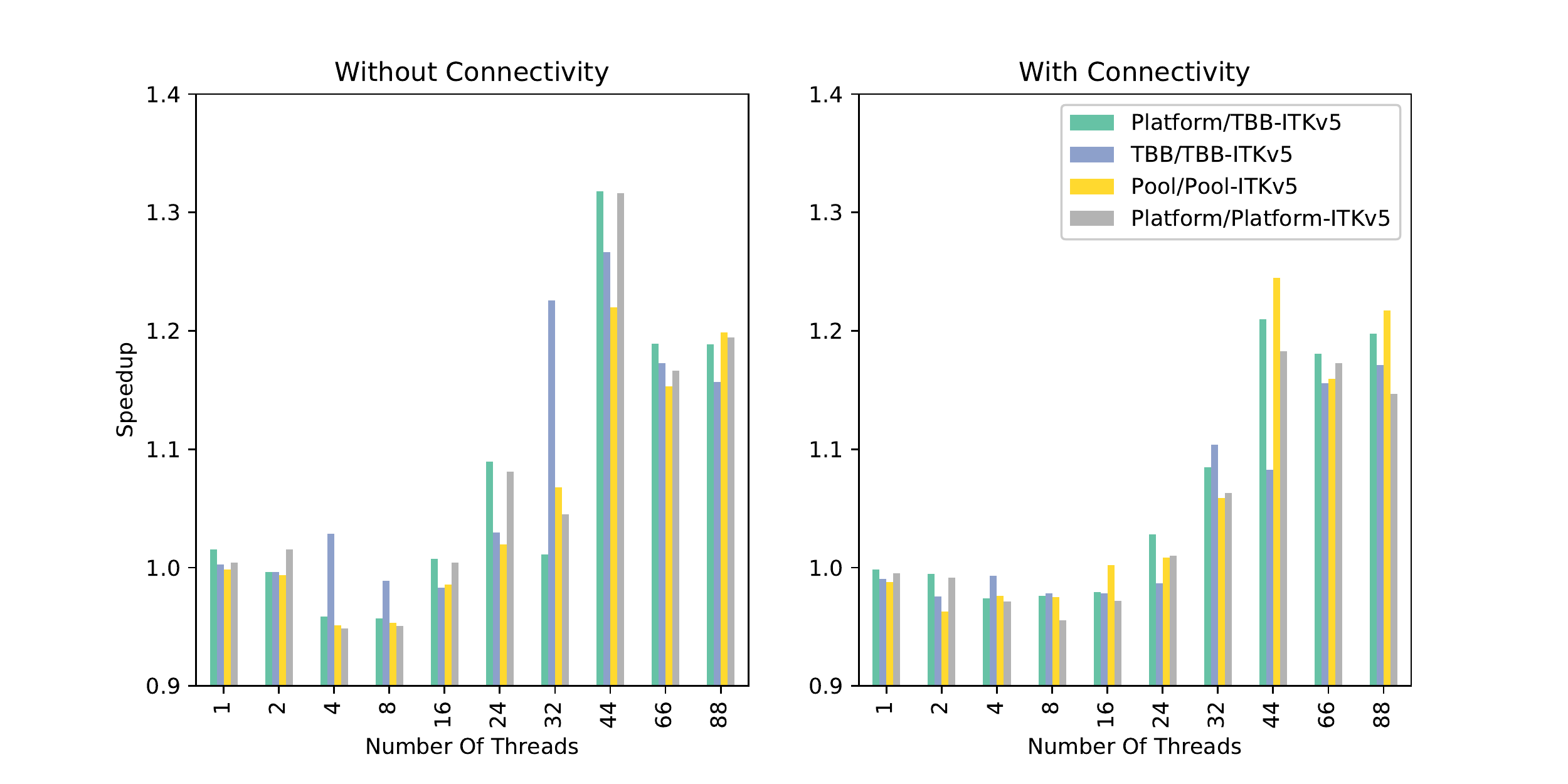}
	\itkcaption[Thread Speedup]{Speedup obtained by
		the ITK SSLIC agorithm wiht the ITKv5 implementation compared the original with a varied number of threads. 
		left: without the connectivity step
		right: with the connectivity step that has a significant single threaded component. }
	\label{fig:ITKv5_speedup}
\end{figure}

The timing results are summarized in Figure \ref{fig:ITKv5_time}. Overall the
performance is quite similar with regard to scalability and efficiency.
However, with four or eight threads the timing is marginally faster for the
original implementation, while with more than 44 threads the ITKv5
implementation has a slight advantage. This is reflected in Figure
\ref{fig:ITKv5_speedup}; we conjecture that the use of a barrier in the original
implementation exhibited poor scalability, while the increase in the number of
dynamic memory allocations is likely responsible for the decrease in performance with the
ITKv5 implementation. Further testing and profiling is required to verify these
conjectures.

\section{Discussion and Conclusion}

In this work we presented SSLIC, an ITK based extension of the SLIC algorithm
that accommodates n-dimensional scalar and multi-channel images and parallelizes
the original sequential implementation. Using a multi-core system we have shown
that our implementation has strong scalability characteristics and is able to
efficiently utilize additional computational resources. When compared to the
SLIC implementation found in the scikit-image toolkit~\cite{scikit-image} we
observed that on a 2D image (Figure~\ref{fig:astronaut}) our single threaded
SSLIC was slower than the scikit-image SLIC, 315ms vs. 166ms, but when using
additional threads it was faster, at 86ms for 4 threads and 52ms for 8 threads.

The closest work to ours is that presented in~\cite{borovec2014} which
describes jSLIC, a SLIC plugin for the ImageJ program. That work described a
parallel implementation of the SLIC algorithm for color images. Beyond the
parallelization, the jSLIC implementation describes a lookup table approach to
conversion from RGB to CIE-Lab color space. As ITK does not explicitly support
the notion of color spaces, both RGB and CIE-Lab images are three channel images.
We assume the image is in CIE-Lab space when using the default weighting
parameter value, otherwise the user needs to set it appropriately or convert the
image to CIE-Lab representation. An additional significant difference is that the
jSLIC algorithm only supports 2D images while SSLIC supports n-dimensional
images. The evaluation of the jSLIC method was carried out on a 4 core machine
with 8Gb RAM, with improved performance when using up to 4 threads. In our case
we observed improved performance even when exceeding the number of physical
cores on our system. Based on the graphs in the jSLIC paper it appears that the
relative efficiency for 2 and 4 threads is approximately 0.71 and 0.45 for an
image of size $[8000 \times 8000]$ while the SSLIC implementation shows better
scalability with relative efficiency for 2 and 4 threads of 0.94 and 0.84 on a
image of size $[2048 \times 1216]$.

We presented a scalable implementation of the SLIC algorithm, SSLIC, a useful
addition to ITK. We demonstrated its performance both qualitatively and
quantitatively on diverse datasets of 2D and 3D, scalar and multi-component, as
well as 2-dimensional and 3-dimensional images.

The SSLIC implementation is available in ITK version 5 in the SuperPixel module (\url{https://itk.org/Doxygen/html/group\textunderscore \textunderscore ITKSuperPixel.html}), in ITK version 4.13.1 it is available in the SimpleITKFilters remote module (\url{https://github.com/SimpleITK/ITKSimpleITKFilters}), and it was originally implemented in a standalone remote module (\url{https://github.com/blowekamp/itkSuperPixel}).

\section*{Acknowledgments}

This work was supported by the Intramural Research Program of the U.S. National Institutes of Health, National
Library of Medicine.

\bibliographystyle{plain}
\bibliography{InsightJournal}

\appendix
	
\newpage
	
\section{Detailed Quantitative Results}

{ \footnotesize
	% 2D
	\captionof{table}{SSLIC performance as a function of number of threads using 2D color (CIE-Lab) image.}
	\label{tab:vh_2d}
	\begin{longtable}{p{1.2cm}| *{6}{p{2cm}}r}
		\hline\\[1mm]
		Number of Threads &   \multicolumn{3}{c}{Without Connectivity} & \multicolumn{3}{c}{With Connectivity}\\
			{} &    Time (sec)                   &  Efficiency  &       Speedup      &    Time (sec)                &  Efficiency                                       &  Speedup                                  \\
	                                  \\
	\hline\\[1mm]
	\endhead
	1                 &               1.36556 &                                   1.00000 &                                1.00000 &            1.47129 &                                1.00000 &                             1.00000 \\
	2                 &               0.70759 &                                   0.96493 &                                1.92987 &            0.78412 &                                0.93818 &                             1.87636 \\
	3                 &               0.48683 &                                   0.93501 &                                2.80503 &            0.55535 &                                0.88310 &                             2.64931 \\
	4                 &               0.36599 &                                   0.93278 &                                3.73112 &            0.43546 &                                0.84467 &                             3.37870 \\
	5                 &               0.30471 &                                   0.89630 &                                4.48152 &            0.36437 &                                0.80759 &                             4.03793 \\
	6                 &               0.26402 &                                   0.86204 &                                5.17222 &            0.31430 &                                0.78019 &                             4.68116 \\
	7                 &               0.22798 &                                   0.85569 &                                5.98984 &            0.26488 &                                0.79351 &                             5.55455 \\
	8                 &               0.20228 &                                   0.84387 &                                6.75099 &            0.24005 &                                0.76613 &                             6.12904 \\
	9                 &               0.18564 &                                   0.81735 &                                7.35614 &            0.21754 &                                0.75148 &                             6.76330 \\
	10                &               0.17500 &                                   0.78030 &                                7.80304 &            0.20157 &                                0.72992 &                             7.29917 \\
	11                &               0.15891 &                                   0.78122 &                                8.59343 &            0.19124 &                                0.69940 &                             7.69337 \\
	12                &               0.15398 &                                   0.73902 &                                8.86822 &            0.18363 &                                0.66769 &                             8.01233 \\
	13                &               0.14088 &                                   0.74561 &                                9.69296 &            0.17173 &                                0.65902 &                             8.56729 \\
	14                &               0.13489 &                                   0.72314 &                               10.12391 &            0.16276 &                                0.64569 &                             9.03963 \\
	15                &               0.12877 &                                   0.70699 &                               10.60492 &            0.16257 &                                0.60334 &                             9.05011 \\
	16                &               0.14586 &                                   0.58514 &                                9.36222 &            0.15151 &                                0.60694 &                             9.71101 \\
	17                &               0.11807 &                                   0.68032 &                               11.56552 &            0.14903 &                                0.58073 &                             9.87243 \\
	18                &               0.11399 &                                   0.66553 &                               11.97957 &            0.13919 &                                0.58723 &                            10.57005 \\
	19                &               0.10692 &                                   0.67222 &                               12.77220 &            0.13443 &                                0.57605 &                            10.94503 \\
	20                &               0.10609 &                                   0.64359 &                               12.87189 &            0.15939 &                                0.46152 &                             9.23045 \\
	21                &               0.09910 &                                   0.65618 &                               13.77981 &            0.14493 &                                0.48340 &                            10.15144 \\
	22                &               0.09583 &                                   0.64770 &                               14.24930 &            0.13295 &                                0.50301 &                            11.06623 \\
	23                &               0.09406 &                                   0.63121 &                               14.51783 &            0.12498 &                                0.51181 &                            11.77172 \\
	24                &               0.09026 &                                   0.63041 &                               15.12986 &            0.12287 &                                0.49894 &                            11.97444 \\
	25                &               0.09029 &                                   0.60495 &                               15.12367 &            0.12446 &                                0.47285 &                            11.82137 \\
	26                &               0.08801 &                                   0.59680 &                               15.51671 &            0.11829 &                                0.47840 &                            12.43828 \\
	27                &               0.08501 &                                   0.59495 &                               16.06374 &            0.14416 &                                0.37800 &                            10.20587 \\
	28                &               0.09119 &                                   0.53482 &                               14.97491 &            0.13670 &                                0.38438 &                            10.76266 \\
	29                &               0.08097 &                                   0.58155 &                               16.86505 &            0.10722 &                                0.47319 &                            13.72264 \\
	30                &               0.08412 &                                   0.54114 &                               16.23411 &            0.10582 &                                0.46346 &                            13.90394 \\
	31                &               0.07887 &                                   0.55849 &                               17.31305 &            0.13351 &                                0.35548 &                            11.01989 \\
	32                &               0.07720 &                                   0.55279 &                               17.68938 &            0.12155 &                                0.37825 &                            12.10400 \\
	33                &               0.09699 &                                   0.42666 &                               14.07986 &            0.12527 &                                0.35589 &                            11.74448 \\
	34                &               0.08110 &                                   0.49522 &                               16.83738 &            0.11108 &                                0.38956 &                            13.24519 \\
	35                &               0.09318 &                                   0.41871 &                               14.65498 &            0.11740 &                                0.35806 &                            12.53204 \\
	36                &               0.07395 &                                   0.51295 &                               18.46609 &            0.12608 &                                0.32414 &                            11.66921 \\
	37                &               0.07362 &                                   0.50131 &                               18.54832 &            0.12421 &                                0.32014 &                            11.84518 \\
	38                &               0.08703 &                                   0.41293 &                               15.69130 &            0.09809 &                                0.39473 &                            14.99982 \\
	39                &               0.08930 &                                   0.39210 &                               15.29205 &            0.12244 &                                0.30812 &                            12.01669 \\
	40                &               0.08479 &                                   0.40264 &                               16.10579 &            0.12145 &                                0.30286 &                            12.11455 \\
	41                &               0.07942 &                                   0.41938 &                               17.19465 &            0.11399 &                                0.31481 &                            12.90719 \\
	42                &               0.09049 &                                   0.35931 &                               15.09095 &            0.12015 &                                0.29157 &                            12.24593 \\
	43                &               0.08762 &                                   0.36243 &                               15.58452 &            0.12035 &                                0.28431 &                            12.22516 \\
	44                &               0.08804 &                                   0.35250 &                               15.51003 &            0.11677 &                                0.28636 &                            12.59977 \\
	45                &               0.08901 &                                   0.34094 &                               15.34222 &            0.11943 &                                0.27377 &                            12.31966 \\
	46                &               0.08561 &                                   0.34677 &                               15.95140 &            0.11862 &                                0.26965 &                            12.40368 \\
	47                &               0.08151 &                                   0.35646 &                               16.75376 &            0.11646 &                                0.26879 &                            12.63310 \\
	48                &               0.08403 &                                   0.33854 &                               16.24996 &            0.11613 &                                0.26394 &                            12.66920 \\
	49                &               0.08603 &                                   0.32394 &                               15.87311 &            0.11563 &                                0.25967 &                            12.72377 \\
	50                &               0.08352 &                                   0.32699 &                               16.34973 &            0.11452 &                                0.25694 &                            12.84689 \\
	51                &               0.08592 &                                   0.31163 &                               15.89289 &            0.11054 &                                0.26097 &                            13.30953 \\
	52                &               0.08289 &                                   0.31683 &                               16.47540 &            0.10923 &                                0.25902 &                            13.46925 \\
	53                &               0.08224 &                                   0.31328 &                               16.60402 &            0.10797 &                                0.25711 &                            13.62694 \\
	54                &               0.07924 &                                   0.31915 &                               17.23413 &            0.11160 &                                0.24415 &                            13.18404 \\
	55                &               0.08319 &                                   0.29847 &                               16.41576 &            0.10735 &                                0.24919 &                            13.70530 \\
	56                &               0.08023 &                                   0.30395 &                               17.02122 &            0.10753 &                                0.24433 &                            13.68245 \\
	57                &               0.08245 &                                   0.29058 &                               16.56311 &            0.10898 &                                0.23686 &                            13.50089 \\
	58                &               0.08110 &                                   0.29032 &                               16.83862 &            0.10819 &                                0.23446 &                            13.59884 \\
	59                &               0.08097 &                                   0.28584 &                               16.86465 &            0.10729 &                                0.23242 &                            13.71306 \\
	60                &               0.08032 &                                   0.28337 &                               17.00237 &            0.10761 &                                0.22788 &                            13.67293 \\
	61                &               0.08145 &                                   0.27484 &                               16.76504 &            0.10308 &                                0.23399 &                            14.27326 \\
	62                &               0.07793 &                                   0.28261 &                               17.52205 &            0.10392 &                                0.22835 &                            14.15763 \\
	63                &               0.07770 &                                   0.27898 &                               17.57576 &            0.10545 &                                0.22147 &                            13.95236 \\
	64                &               0.07504 &                                   0.28433 &                               18.19733 &            0.10084 &                                0.22796 &                            14.58976 \\
	65                &               0.07609 &                                   0.27610 &                               17.94667 &            0.10309 &                                0.21957 &                            14.27200 \\
	66                &               0.07713 &                                   0.26826 &                               17.70518 &            0.10714 &                                0.20808 &                            13.73302 \\
	67                &               0.07851 &                                   0.25960 &                               17.39349 &            0.10329 &                                0.21260 &                            14.24436 \\
	68                &               0.08251 &                                   0.24340 &                               16.55091 &            0.10417 &                                0.20770 &                            14.12380 \\
	69                &               0.07721 &                                   0.25631 &                               17.68544 &            0.10247 &                                0.20809 &                            14.35821 \\
	70                &               0.08099 &                                   0.24088 &                               16.86133 &            0.10283 &                                0.20439 &                            14.30728 \\
	71                &               0.08168 &                                   0.23548 &                               16.71889 &            0.10971 &                                0.18888 &                            13.41070 \\
	72                &               0.07975 &                                   0.23783 &                               17.12397 &            0.10666 &                                0.19159 &                            13.79482 \\
	73                &               0.07844 &                                   0.23849 &                               17.40967 &            0.10800 &                                0.18661 &                            13.62279 \\
	74                &               0.08142 &                                   0.22666 &                               16.77269 &            0.10480 &                                0.18971 &                            14.03859 \\
	75                &               0.07734 &                                   0.23542 &                               17.65644 &            0.10729 &                                0.18285 &                            13.71358 \\
	76                &               0.08071 &                                   0.22262 &                               16.91895 &            0.10658 &                                0.18164 &                            13.80429 \\
	77                &               0.07951 &                                   0.22305 &                               17.17516 &            0.10967 &                                0.17422 &                            13.41499 \\
	78                &               0.07772 &                                   0.22525 &                               17.56983 &            0.10318 &                                0.18282 &                            14.25974 \\
	79                &               0.08092 &                                   0.21360 &                               16.87464 &            0.10740 &                                0.17340 &                            13.69888 \\
	80                &               0.08219 &                                   0.20767 &                               16.61375 &            0.10658 &                                0.17256 &                            13.80519 \\
	81                &               0.07570 &                                   0.22271 &                               18.03914 &            0.10385 &                                0.17490 &                            14.16703 \\
	82                &               0.07456 &                                   0.22334 &                               18.31375 &            0.10881 &                                0.16490 &                            13.52187 \\
	83                &               0.07780 &                                   0.21148 &                               17.55266 &            0.10587 &                                0.16744 &                            13.89762 \\
	84                &               0.07943 &                                   0.20467 &                               17.19248 &            0.10997 &                                0.15927 &                            13.37863 \\
	85                &               0.07897 &                                   0.20344 &                               17.29261 &            0.10550 &                                0.16407 &                            13.94636 \\
	86                &               0.07874 &                                   0.20165 &                               17.34225 &            0.10876 &                                0.15730 &                            13.52771 \\
	87                &               0.08251 &                                   0.19024 &                               16.55091 &            0.10380 &                                0.16292 &                            14.17412 \\
	88                &               0.08240 &                                   0.18833 &                               16.57279 &            0.10247 &                                0.16317 &                            14.35864 \\
	\hline	
	\end{longtable}
	
}

\newpage

\addtocounter{table}{-1}
{ \footnotesize
	% 3D
	\captionof{table}{SSLIC performance as a function of number of threads using 3D color (CIE-Lab) image.}
	\label{tab:vh_3d}
	\begin{longtable}{p{1.2cm}| *{6}{p{2cm}}r}
		\hline\\[1mm]
		Number of Threads &   \multicolumn{3}{c}{Without Connectivity}& \multicolumn{3}{c}{With Connectivity}\\
		{} &    Time (sec)                   &  Efficiency  &       Speedup      &    Time (sec)                &  Efficiency                                       &  Speedup                                  \\
		\hline\\[1mm]
		1                 &              7609.695 &                                     1.000 &                                  1.000 &           8408.861 &                                  1.000 &                               1.000 \\
		2                 &              3798.146 &                                     1.002 &                                  2.004 &           4188.603 &                                  1.004 &                               2.008 \\
		4                 &              2043.198 &                                     0.931 &                                  3.724 &           2322.334 &                                  0.905 &                               3.621 \\
		8                 &              1094.085 &                                     0.869 &                                  6.955 &           1268.391 &                                  0.829 &                               6.630 \\
		16                &               640.219 &                                     0.743 &                                 11.886 &            777.877 &                                  0.676 &                              10.810 \\
		24                &               462.156 &                                     0.686 &                                 16.466 &            568.193 &                                  0.617 &                              14.799 \\
		32                &               352.315 &                                     0.675 &                                 21.599 &            455.163 &                                  0.577 &                              18.474 \\
		44                &               263.890 &                                     0.655 &                                 28.837 &            364.519 &                                  0.524 &                              23.068 \\
		66                &               244.558 &                                     0.471 &                                 31.116 &            339.879 &                                  0.375 &                              24.741 \\
		88                &               229.544 &                                     0.377 &                                 33.151 &            328.298 &                                  0.291 &                              25.614 \\
		\hline
	\end{longtable}
}

\addtocounter{table}{-1}
{ \footnotesize
    % Without Connectivity Timing
    \captionof{table}{SSLIC execution time as a function of number of threads using 2D color (CIE-Lab) image for the classic ITK threading models and then modern ITKv5 for the Platform, Pool and Intel TBB multi-threader backends with the connectivity step disabled.}
	\label{tab:ITKv5_time_wo}
    \begin{longtable}{p{1.2cm}| *{6}{p{2cm}}r}
		\hline\\[1mm]
		Number Of Threads &  {} & {} & {} & {} & {} & \\
		{} & Platform-ITKv5 &  Platform &  Pool-ITKv5 &    Pool &  TBB-ITKv5 &     TBB \\
		\hline\\[1mm]
		1                 &         1.34867 &   1.35434 &     1.33995 & 1.33794 &    1.33399 & 1.33770 \\
		2                 &         0.68490 &   0.69555 &     0.69927 & 0.69472 &    0.69832 & 0.69568 \\
		4                 &         0.38042 &   0.36096 &     0.37429 & 0.35614 &    0.37650 & 0.38728 \\
		8                 &         0.21055 &   0.20014 &     0.20832 & 0.19859 &    0.20917 & 0.20680 \\
		16                &         0.12448 &   0.12499 &     0.12277 & 0.12098 &    0.12407 & 0.12199 \\
		24                &         0.08972 &   0.09700 &     0.08750 & 0.08921 &    0.08905 & 0.09168 \\
		32                &         0.07299 &   0.07627 &     0.07103 & 0.07583 &    0.07541 & 0.09243 \\
		44                &         0.06716 &   0.08840 &     0.06633 & 0.08090 &    0.06707 & 0.08494 \\
		66                &         0.06681 &   0.07791 &     0.06489 & 0.07481 &    0.06552 & 0.07683 \\
		88                &         0.06612 &   0.07897 &     0.06520 & 0.07814 &    0.06646 & 0.07689 \\
	\hline
	\end{longtable}
}

\newpage

\addtocounter{table}{-1}
{ \footnotesize
	% With Connectivity Timing
	\captionof{table}{SSLIC execution time as a function of number of threads using 2D color (CIE-Lab) image for the classic ITK threading models and then modern ITKv5 for the Platform, Pool and Intel TBB multi-threader backends with the connectivity step enabled.}
	\label{tab:ITKv5_time_w}
	\begin{longtable}{p{1.2cm}| *{6}{p{2cm}}r}
		\hline\\[1mm]
		Number Of Threads &  {} & {} & {} & {} & {} & \\
		{} & Platform-ITKv5 &  Platform &  Pool-ITKv5 &    Pool &  TBB-ITKv5 &     TBB \\
		\hline\\[1mm]
		1                 &         1.49384 &   1.48678 &     1.49108 & 1.47300 &    1.48905 & 1.47446 \\
		2                 &         0.78172 &   0.77497 &     0.79180 & 0.76258 &    0.77926 & 0.76014 \\
		4                 &         0.42666 &   0.41437 &     0.42546 & 0.41542 &    0.42530 & 0.42242 \\
		8                 &         0.24895 &   0.23785 &     0.24425 & 0.23820 &    0.24367 & 0.23841 \\
		16                &         0.15398 &   0.14963 &     0.15049 & 0.15083 &    0.15281 & 0.14950 \\
		24                &         0.11750 &   0.11866 &     0.11588 & 0.11687 &    0.11543 & 0.11390 \\
		32                &         0.09789 &   0.10405 &     0.09599 & 0.10164 &    0.09592 & 0.10590 \\
		44                &         0.09570 &   0.11320 &     0.09330 & 0.11615 &    0.09358 & 0.10134 \\
		66                &         0.09016 &   0.10575 &     0.08906 & 0.10324 &    0.08958 & 0.10352 \\
		88                &         0.09178 &   0.10528 &     0.08576 & 0.10437 &    0.08789 & 0.10294 \\
	\hline
	\end{longtable}
}

\addtocounter{table}{-1}
{ \footnotesize
	% Without Connectivity ITKv5 Speedup
	\captionof{table}{The speed up of the modern ITKv5 implementation of SSLIC compared to the original implementation as a function of number of threads using 2D color (CIE-Lab) image with the connectivity step disabled.}
	\label{tab:ITKv5_speedup_wo}
	\begin{longtable}{p{1.2cm}| *{4}{p{2cm}}r}
		\hline\\[1mm]
		Number Of Threads &  {} & {} & {} & {} & \\
		{} &  Platform/TBB-ITKv5 &  TBB/TBB-ITKv5 &  Pool/Pool-ITKv5 &  Platform/Platform-ITKv5 \\
		\hline\\[1mm]
		1                 &             1.01525 &        1.00278 &          0.99850 &                  1.00421 \\
		2                 &             0.99604 &        0.99622 &          0.99350 &                  1.01556 \\
		4                 &             0.95873 &        1.02863 &          0.95149 &                  0.94886 \\
		8                 &             0.95683 &        0.98869 &          0.95332 &                  0.95058 \\
		16                &             1.00746 &        0.98329 &          0.98545 &                  1.00409 \\
		24                &             1.08933 &        1.02956 &          1.01952 &                  1.08117 \\
		32                &             1.01131 &        1.22566 &          1.06760 &                  1.04488 \\
		44                &             1.31797 &        1.26646 &          1.21963 &                  1.31622 \\
		66                &             1.18914 &        1.17266 &          1.15290 &                  1.16616 \\
		88                &             1.18835 &        1.15696 &          1.19843 &                  1.19447 \\
	\hline
	\end{longtable}
}

\newpage

\addtocounter{table}{-1}
{ \footnotesize
	% With Connectivity ITKv5 Speedup
	\captionof{table}{The speed up of the modern ITKv5 implementation of SSLIC compared to the original implementation as a function of number of threads using 2D color (CIE-Lab) image with the connectivity step enabled.}
	\label{tab:ITKv5_speedup_w}
	\begin{longtable}{p{1.2cm}| *{4}{p{2cm}}r}
		\hline\\[1mm]
		Number Of Threads &  {} & {} & {} & {} & \\
		{} &  Platform/TBB-ITKv5 &  TBB/TBB-ITKv5 &  Pool/Pool-ITKv5 &  Platform/Platform-ITKv5 \\
		\hline\\[1mm]
		1                 &             0.99847 &        0.99020 &          0.98787 &                  0.99527 \\
		2                 &             0.99450 &        0.97547 &          0.96311 &                  0.99137 \\
		4                 &             0.97430 &        0.99322 &          0.97639 &                  0.97120 \\
		8                 &             0.97612 &        0.97841 &          0.97524 &                  0.95542 \\
		16                &             0.97921 &        0.97837 &          1.00227 &                  0.97178 \\
		24                &             1.02801 &        0.98679 &          1.00853 &                  1.00989 \\
		32                &             1.08472 &        1.10396 &          1.05886 &                  1.06297 \\
		44                &             1.20971 &        1.08289 &          1.24491 &                  1.18298 \\
		66                &             1.18049 &        1.15558 &          1.15928 &                  1.17297 \\
		88                &             1.19775 &        1.17124 &          1.21704 &                  1.14699 \\

	\hline
\end{longtable}
}
	
\end{document}